\documentclass[final, custom]{anthology-ch}

\newif\ifanonymous
\anonymousfalse

\usepackage{tikz}
\usetikzlibrary{calc, arrows.meta, shapes.geometric}
\usepackage{placeins}
\usepackage{xurl} % line breaks anywhere in URLs (footnotes, availability section)
\usepackage{fvextra} % listings of the appendix, with line breaks inside long strings
\AtBeginDocument{\appto\biburlsetup{\Urlmuskip=0mu\relax}} % no stretched URLs in the references
\usepackage[protrusion=false, expansion=false, tracking=true, kerning=false, spacing=false]{microtype}
\input{glyphtounicode}
\makeatletter
\let\ciae@unicodechar\DeclareUnicodeCharacter
\let\ciae@letters\@empty
\def\ciae@armenian#1#2#3\@nil#4#5{%
    \ifnum\pdfstrcmp{#1#2}{05}=0
        \ciae@unicodechar{#4}{#5}%
        \g@addto@macro\ciae@letters{\ciae@letter{#4}#5}%
    \fi}
\renewcommand{\DeclareUnicodeCharacter}[2]{\ciae@armenian#1\@nil{#1}{#2}}
\usepackage[latin, safe]{armtex}
\AddToHook{begindocument/end}{\let\DeclareUnicodeCharacter\ciae@unicodechar}
\newcommand{\ciae@letter}[2]{%
    \DeclareTextSymbolDefault{#2}{OT6}%
    \pdfglyphtounicode{\expandafter\@gobble\string#2}{#1}}
\ciae@letters
\ciae@unicodechar{23A1}{\ensuremath{\lceil}}
\ciae@unicodechar{23A4}{\ensuremath{\rceil}}
\ciae@unicodechar{23A3}{\ensuremath{\lfloor}}
\ciae@unicodechar{23A6}{\ensuremath{\rfloor}}
\ciae@unicodechar{02BB}{\textquoteleft}
\ciae@unicodechar{1E59}{\.r}
\makeatother
\DeclareFontFamily{OT6}{lmtt}{\hyphenchar\font=-1}
\DeclareFontShape{OT6}{lmtt}{m}{n}{<-> ssub * cmss/m/n}{}
\pdfglyphtounicode{ceilingleft}{23A1}
\pdfglyphtounicode{ceilingright}{23A4}
\pdfglyphtounicode{floorleft}{23A3}
\pdfglyphtounicode{floorright}{23A6}
\newcommand{\raleway}{\fontfamily{Raleway-TLF}\fontseries{m}\fontshape{n}\selectfont}
\newcommand{\ralewaysemibold}{\fontfamily{Raleway-TLF}\fontseries{sb}\fontshape{n}\selectfont}
\newcommand{\hy}[1]{{\fontencoding{OT6}\fontfamily{cmss}\selectfont #1}}
\newcommand{\sig}[1]{#1}

\definecolor{ciaeNuit}{HTML}{163A54}
\definecolor{ciaeRubrique}{HTML}{8A2A1E}
\definecolor{ciaeEncreDouce}{HTML}{45423D}
\definecolor{ciaeFilet}{HTML}{D9D5CD}

\newcommand{\kicker}[1]{{\ralewaysemibold\textls[80]{\MakeUppercase{#1}}}}

\newcommand{\boitetexte}[2]{%
    \hyphenpenalty=10000\exhyphenpenalty=10000\relax
    {\color{ciaeNuit}\ralewaysemibold\fontsize{8}{9.5}\selectfont #1\par}%
    \if\relax\detokenize{#2}\relax\else
        \vspace{1.5pt}{\color{ciaeEncreDouce}\raleway\fontsize{6.6}{8}\selectfont #2\par}%
    \fi}

\newcommand{\callout}[3]{%
    \node[circle, fill=ciaeRubrique, text=white, inner sep=0pt, minimum size=0.42cm,
          font=\ralewaysemibold\fontsize{8}{8}\selectfont, draw=white, line width=0.6pt]
        at (#1, #2) {#3};}

\title{Towards an Armenian Epigraphic Corpus}

\author[1]{Chahan Vidal-Gorène}[orcid=0000-0003-1567-6508]
\author[2]{Arpine Avetisyan}[orcid=0000-0002-6436-866X]
\author[1]{Edita Matevosyan}[orcid=]
\author[1]{Seda Kirakosyan}[orcid=]
\author[3,4]{Victoria Khurshudyan}[orcid=0000-0002-9960-2516]
\author[2,5]{Pollet Samvelian}[orcid=0000-0002-6622-4117]

\affiliation{1}{Calfa, Paris, France}
\affiliation{2}{Université Sorbonne Nouvelle, Paris, France}
\affiliation{3}{SeDyL (UMR 8202), Villejuif, France}
\affiliation{4}{Inalco, Paris, France}
\affiliation{5}{Lattice, Paris, France}

\keywords[eng]{Armenian epigraphy, OCR, endangered heritage, vision-language models}

\pubyear{2027}
\pubvolume{1}
\pagestart{1}
\pageend{1}
\paperorder{1}
\conferencename{DH2027}
\conferenceeditors{}
\doi{00000/00000}
\customhead{} % no running head for now

\ifanonymous
  \AtBeginDocument{\hypersetup{pdfauthor={Anonymous}}}
\fi

\newcommand{\lienfn}[1]{\footnote{\ifanonymous Link withheld for review.\else\url{#1}\fi}}

\begin{document}

\maketitle

\begin{abstract}
The \emph{Divan hay vimagrutʻyan} (1960--2017), the printed corpus of Armenian inscriptions, comprises
nearly 7,600 records; a growing number of the inscriptions they describe, damaged or
destroyed in recent decades, survive only in its pages. Armenian epigraphy nevertheless lacks a
digital resource of the kind available for Greek, Latin or Georgian. This paper presents the \emph{Corpus
Inscriptionum Armenicarum Electronicum}, which turns the ten volumes of the \emph{Divan} into an open
digital corpus. Layout detection, OCR and a vision-language model produce a first structured reading
of each record, which specialists then check on a proofreading platform. First corrections show that
letters are read reliably and that residual errors concentrate in the editorial signs. The corpus is designed as a lasting
platform for documenting Armenian inscriptions and as a source of silver data in Classical and
Middle Armenian for language models and computational linguistics.
\end{abstract}

% ============================================================================
\section{Introduction}
% ============================================================================

Armenian inscriptions have been used as historical sources since the Middle Ages, by
Stepʻanos Orbelean among others. In the nineteenth century, Catholicos Nersēs Ashtaraketsʻi had
them collected to save them from loss, and scholars such as Sargis Jalaleantsʻ, Ghevond Alishan and
Makar Barkhudareantsʻ recorded many more, but without a common method
\cite{orbeli1966,greenwood2014}.
The scientific collection of Armenian inscriptions began with the excavations of Ani, where N. Marr
and H. Orbeli set its editorial principles: every inscription was to be published with the
utmost accuracy, whatever its size or state of preservation. A permanent epigraphic expedition of the Academy of
Sciences of Armenia later applied them, with the aim of documenting every Armenian inscription
\cite{orbeli1966}. Since 1960,
the Institute of Archaeology and Ethnography of the National Academy of Sciences of Armenia has
published the results as the \emph{Divan hay vimagrutʻyan} (\hy{Դիվան հայ վիմագրության}, Corpus of
Armenian Inscriptions)\footnote{Scans of the ten volumes are available from the Fundamental
Scientific Library of the National Academy of Sciences of Armenia; volume I:
\url{http://serials.flib.sci.am/openreader/vimagrutyun_1/book/content.html}.}: ten volumes, organised by region, whose nearly 7,600 records describe inscriptions
from the fifth to the eighteenth century \cite{divan}. For Orbeli, these texts are a primary source
on the secular, economic and social life that historiography largely ignores \cite{orbeli1966}.

Each record gives the location of the inscription, a description of its position and state of
preservation, its previous editions and a diplomatic transcription, in which conventional signs
mark lacunae, the editor's restorations and line divisions. The editors chose to document letter
forms and the several hundred ligatures of Armenian inscriptions by facsimile drawings, placed beside
the typeset transcription.

Armenian inscriptions themselves are at risk. By the 1960s, some previously recorded inscriptions had
already disappeared \cite{barkhudaryan1982}; weathering, neglect, war and deliberate destruction have
since damaged or erased whole sites, from the cross-stones of Julfa \cite{petrosyan2020culture} to the churches and cemeteries of Artsakh
(Nagorno-Karabakh), whose destruction satellite imagery has documented since 2021
\cite{lindsay2024}. For a growing number of inscriptions, the \emph{Divan} is now the only surviving
witness, yet, beyond printed thematic corpora \cite{greenwood2004,greenwood2015}, Armenian epigraphy has no digital
corpus of its texts, whereas Greek and Latin epigraphy increasingly rests on open digital
corpora \cite{bodel2024} and a digital corpus of Georgian inscriptions\footnote{Epigraphic
Corpus of Georgia: \url{https://epigraphy.iliauni.edu.ge}; on Old Georgian inscriptions, see also
\textcite{hoenen2016}.} has existed since 2015. Systematic 3D documentation of the monuments now
complements these records: the Armenian Cultural Heritage
Institute\footnote{\url{https://armenianculturalheritageinstitute.org/en}}, which grew out of a
scanning project launched in 2020, has recorded hundreds of sites in Armenia and Artsakh by
photogrammetry and 3D scanning, and makes the models freely accessible \cite{achi}.

The \emph{Corpus Inscriptionum Armenicarum Electronicum} (CIAE)\lienfn{https://epigraphy.calfa.fr}
turns the \emph{Divan} into an open digital corpus with three aims: to make the printed volumes
searchable; to provide a lasting platform on which the records of Armenian epigraphic heritage can
be corrected and enriched; and to supply silver data in Classical and Middle Armenian for
computational linguistics, the digital humanities and language models, for instance to decipher
damaged inscriptions.

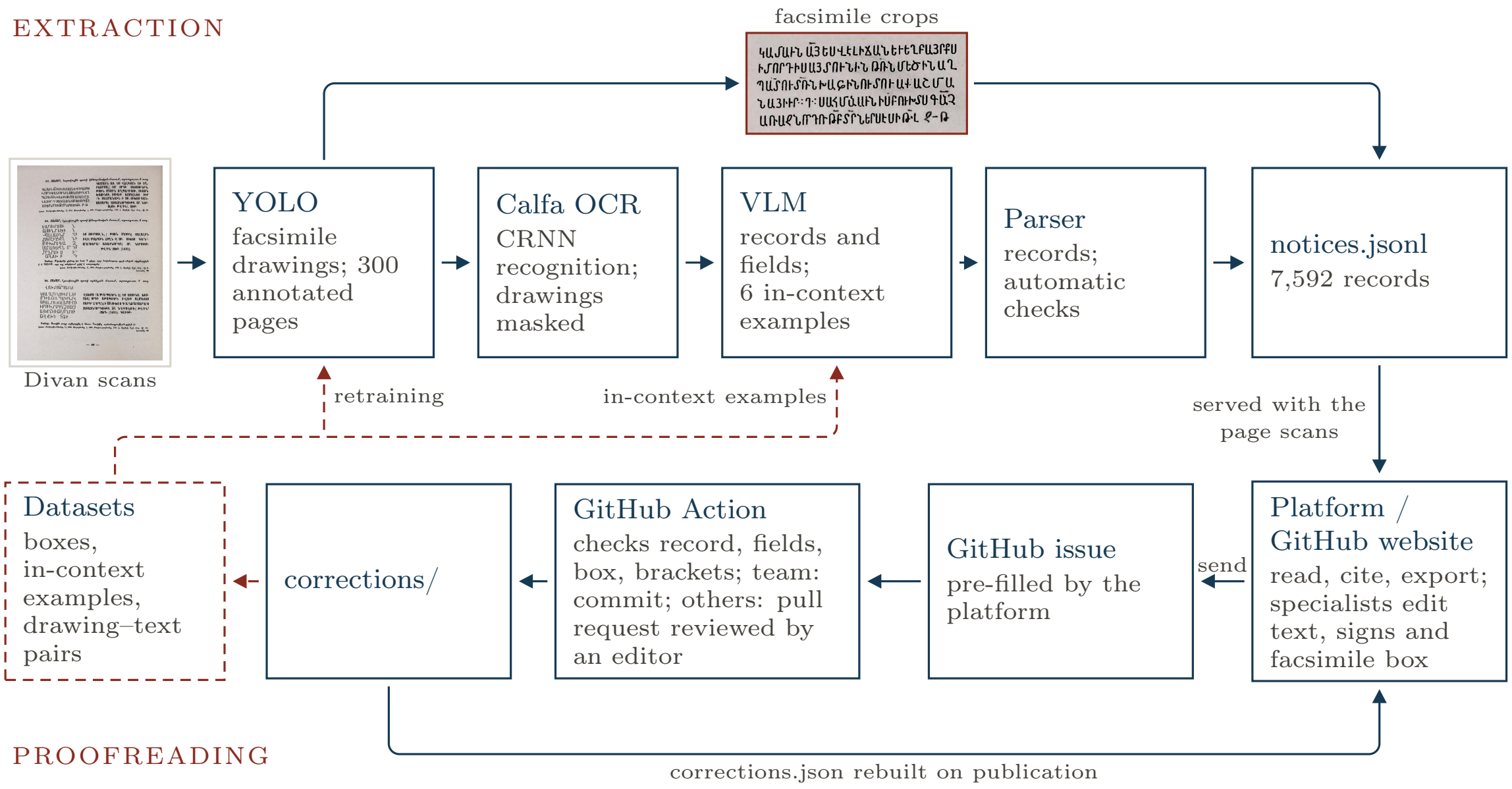
\begin{figure}[t]
\centering
\begin{tikzpicture}[
        font=\raleway,
        boite/.style={draw=ciaeNuit, line width=0.6pt, fill=white,
                      minimum height=1.5cm, inner xsep=5pt, inner ysep=4pt, align=flush left},
        lien/.style={-{Triangle[length=4pt, width=4pt]}, line width=0.6pt, draw=ciaeNuit,
                     rounded corners=3pt, shorten <=2pt, shorten >=2pt},
        retour/.style={lien, draw=ciaeRubrique, dash pattern=on 3pt off 2pt},
        etiq/.style={font=\raleway\fontsize{6.5}{7.8}\selectfont, text=ciaeEncreDouce,
                     inner sep=1.5pt, align=center}]

    % --- Extraction --------------------------------------------------------------
    \node[anchor=north west, inner sep=0pt, text=ciaeRubrique] at (0, 2.45)
        {\fontsize{7}{7}\selectfont\kicker{Extraction}};

        \node[inner sep=0pt] (page) at (0.8, 0) {\includegraphics[height=2.07cm]{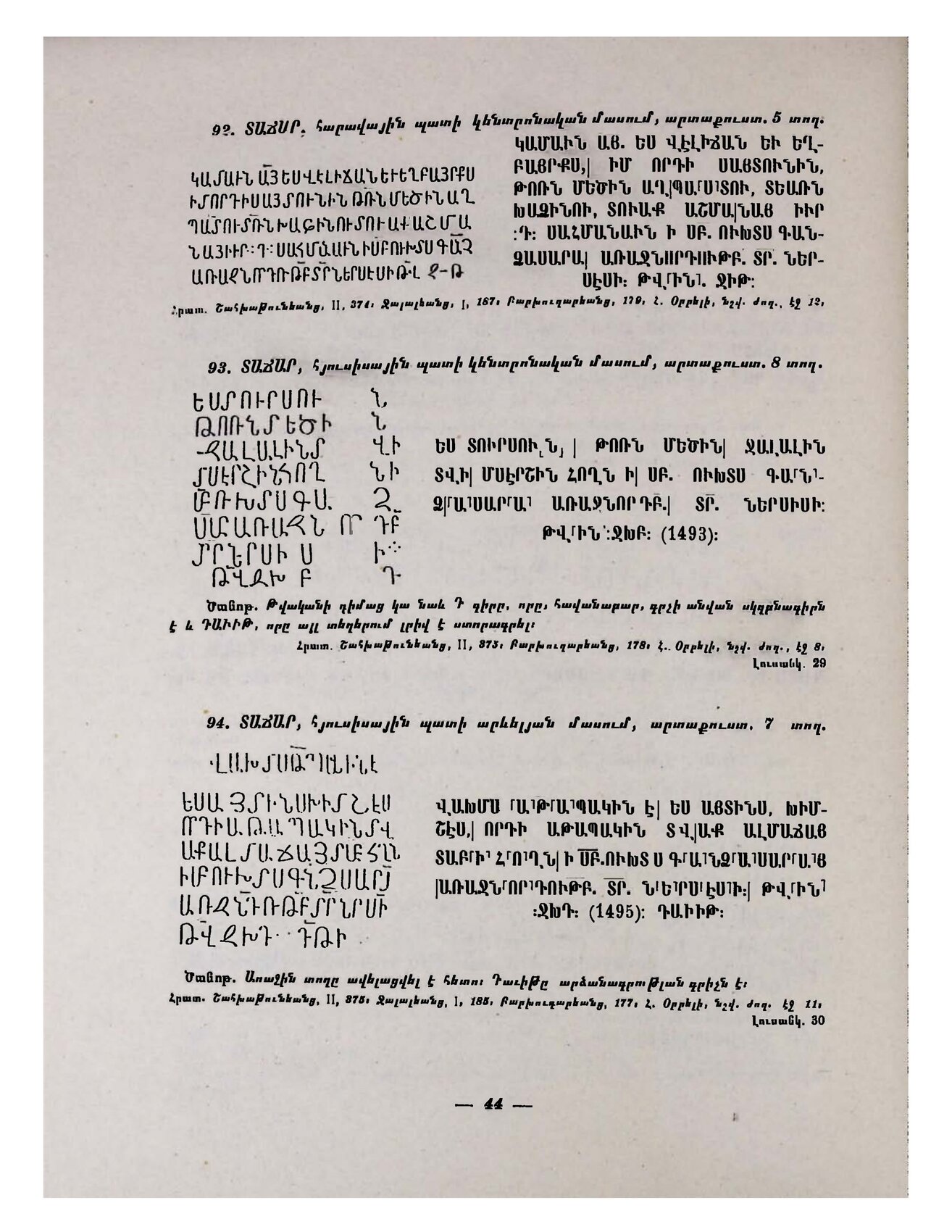}};
    \draw[ciaeFilet, line width=0.6pt] (page.south west) rectangle (page.north east);
    \node[etiq, anchor=north] at (page.south) {\emph{Divan} scans};

        \node[boite, minimum width=2.2cm, text width=1.85cm, minimum height=1.9cm] (yolo) at (3.15, 0)
        {\boitetexte{YOLO}{facsimile drawings; 300 annotated pages}};
        \node[boite, minimum width=2.0cm, text width=1.65cm, minimum height=1.9cm] (ocr) at (5.7, 0)
        {\boitetexte{Calfa OCR}{CRNN recognition; drawings masked}};
        \node[boite, minimum width=2.3cm, text width=1.95cm, minimum height=1.9cm] (vlm) at (8.3, 0)
        {\boitetexte{VLM}{records and fields; 6~in-context examples}};
    \node[boite, minimum width=2.2cm, text width=1.85cm, minimum height=1.9cm] (parse) at (10.9, 0)
        {\boitetexte{Parser}{records; automatic checks}};
        \node[boite, minimum width=2.5cm, text width=2.2cm, minimum height=1.9cm] (db) at (13.75, 0)
        {\boitetexte{notices.jsonl}{7,592 records}};

    \node[inner sep=0pt] (crop) at (8.5, 1.8) {\includegraphics[width=2.2cm]{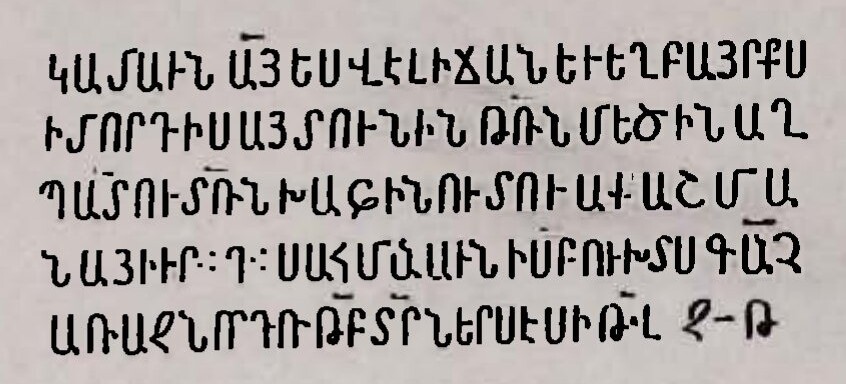}};
    \draw[ciaeRubrique, line width=0.6pt] (crop.south west) rectangle (crop.north east);
    \node[etiq, anchor=south] at (crop.north) {facsimile crops};

    \draw[lien] (page.east |- yolo.west) -- (yolo.west);
    \draw[lien] (yolo.east) -- (ocr.west);
    \draw[lien] (ocr.east) -- (vlm.west);
        \draw[lien] (vlm.east) -- (parse.west);
    \draw[lien] (parse.east) -- (db.west);
    \draw[lien] (yolo.north) |- (crop.west);
    \draw[lien] (crop.east) -| (db.north);

    % --- Proofreading ------------------------------------------------------------
    \node[boite, minimum width=2.5cm, text width=2.2cm, minimum height=1.95cm] (site) at (13.75, -3.2)
        {\boitetexte{Platform /\\GitHub website}{read, cite, export; specialists edit text, signs
         and facsimile box}};
    \node[boite, minimum width=2.6cm, text width=2.3cm, minimum height=1.95cm] (issue) at (10.55, -3.2)
        {\boitetexte{GitHub issue}{pre-filled by the platform}};
    \node[boite, minimum width=3.0cm, text width=2.7cm, minimum height=1.95cm] (action) at (7.0, -3.2)
        {\boitetexte{GitHub Action}{checks record, fields, box, brackets; team: commit;
         others: pull request reviewed by an editor}};
    \node[boite, minimum width=2.4cm, text width=2.1cm, minimum height=1.95cm] (corr) at (3.8, -3.2)
        {\boitetexte{corrections/}{}};
    \node[boite, draw=ciaeRubrique, dash pattern=on 3pt off 2pt, minimum width=2.1cm,
          text width=1.85cm, minimum height=1.95cm] (train) at (1.05, -3.2)
        {\boitetexte{Datasets}{boxes, in-context examples, drawing–text pairs}};

    \draw[lien] (db.south) -- node[etiq, left=2pt] {served with the\\page scans} (site.north);
    \draw[lien] (site.west) -- node[etiq, above=1pt] {send} (issue.east);
    \draw[lien] (issue.west) -- (action.east);
    \draw[lien] (action.west) -- (corr.east);
    \draw[lien] (corr.south) -- ++(0, -0.75) -| node[etiq, pos=0.25, below=1pt]
        {corrections.json rebuilt on publication} (site.south);
    \draw[retour] (corr.west) -- (train.east);
    \draw[retour] (train.north) -- (train.north |- 0, -1.75) -- (vlm.south |- 0, -1.75)
        -- node[etiq, left=1pt] {in-context examples} (vlm.south);
    \draw[retour] (yolo.south |- 0, -1.75) -- node[etiq, right=1pt] {retraining} (yolo.south);

    \node[anchor=west, inner sep=0pt, text=ciaeRubrique] at (0, -4.95)
        {\fontsize{7}{7}\selectfont\kicker{Proofreading}};
\end{tikzpicture}
\caption{Extraction (top) and proofreading (bottom). Corrections are stored alongside
the machine output; corrected boxes and transcriptions feed back into the detector, the in-context examples of the
VLM and training data for reading models (dashed).}
\label{fig:pipeline}
\end{figure}

% ============================================================================
\section{Related work in computational epigraphy}
% ============================================================================

Deep learning now assists epigraphers with damaged texts \cite{sommerschield2023}: Ithaca restores
missing characters in Greek inscriptions and attributes them to a region and a period, raising
historians' restoration accuracy from 25 to 72\,\% \cite{assael2022},\footnote{Ithaca is trained on I.PHI, 78,608
inscriptions derived from the Searchable Greek Inscriptions of the Packard Humanities Institute
\cite{phi}.} and Aeneas extends the approach to Latin and retrieves textual parallels
\cite{assael2025}.\footnote{Aeneas is trained on 176,861 inscriptions from the Epigraphic Database
Roma, the Epigraphic Database Heidelberg and the Epigraphik-Datenbank Clauss/Slaby
\cite{edr,edh,edcs}, only 5\,\% of them paired with images. Both models have since been coupled with
a large language model, Gemini, which historians can query in natural language to attribute, restore
and analyse inscriptions:
\url{https://deepmind.google/science/workflows/conversing-with-antiquity/}.} \textcite{zottin2026gaps} pursue Ithaca's restoration task in a low-resource setting: they
recast lacuna filling as a constrained visual-linguistic inference based on character n-gram
statistics, which sharply reduces the number of candidate restorations that epigraphists must
inspect.

Image analysis has also been applied to inscriptions, and Armenian has already been the subject of
experiments: \textcite{vidalgorene2024} detect inscriptions, characters and lines in 150 images of
damaged inscriptions and graffiti (7,294 characters)
using YOLOv8, with a precision of 0.91 and a recall of 0.88 for localising inscriptions and of
0.91 and 0.84 for detecting characters; a Siamese network with a Vision Transformer backbone then
identifies characters by similarity (precision 0.67, recall 0.71). The small size of the dataset and
the severe damage to the inscriptions remain the main limitations, but the results are promising for
better-preserved inscriptions. A new strategy was introduced at ICDAR 2026 with EpiSAM \cite{sharma2026episam}, which extends
the promptable segmentation of Segment Anything (SAM) \cite{kirillov2023sam} to inscriptions:
prompted on a character, the model also predicts its neighbours so as to resolve the boundaries
between them. Applied to 148 images of degraded Indic stone inscriptions (13,498 characters), it
reaches a per-character IoU of 66.5\,\% against 60.0\,\% for the original SAM, and transfers without
retraining to unseen scripts such as Roman, Brahmi and Thai. Other recent work includes a
multi-task benchmark for Khmer inscriptions \cite{thuon2026angkorian} and the recognition of the
medieval graffiti of Saint Sophia Cathedral, Kyiv \cite{karimi2026graffiti}.

More broadly, multimodal language models now rival dedicated OCR on historical documents
\cite{humphries2025,greif2025}. They remain prone to hallucination \cite{greif2025,vidalgorene2026paris},
to the reproduction of in-context examples and to the loss of small glyphs, but constrained tasks
guided by in-context learning have proved effective on sources in under-resourced languages and
scripts, from the Western Armenian press \cite{vidalgorene2026paris} to Arabic-script handwriting,
where they reach competitive character error rates and provide first-pass annotations
\cite{vidalgorene2026amidda}.

To date, no Armenian epigraphic corpus comparable in scale to those behind Ithaca and Aeneas, and no
model of this kind, is available. Such a corpus is a prerequisite for computational approaches of
the kind presented above, and is also important for preserving this epigraphic heritage. Projects
addressing this need are under way~\cite{tamrazyan2024guardianship}, notably on the encoding of Armenian inscriptions in EpiDoc
\cite{epidoc}: \textcite{tamrazyan2026heritage} map Armenian editorial conventions onto the Leiden
and EpiDoc frameworks. The same need motivates the present work.

% ============================================================================
\section{Methodology and results}
% ============================================================================

\subsection{Extraction pipeline}
\label{sec:pipeline}

The pipeline (Figure~\ref{fig:pipeline}) combines three models. (i)~A YOLO11n object detection
model, trained on 300 page images annotated with a single class, the facsimile drawing, locates the
drawings, which are cropped for display on the platform and masked on the page so that they do not
add noise to the OCR output; plates and blank sheets are discarded at this stage.
(ii)~Calfa OCR, a CRNN recogniser \cite{vidalgorene2021}, then reads the printed text: specialised
CRNNs remain more accurate than VLMs \cite{vidalgorene2026paris,vidalgorene2026amidda}, and the
recogniser limits the hallucinations of the VLM, leaving it to structure the page, a task that would
otherwise require a separate layout model. (iii)~A vision-language model
(VLM),\footnote{The pipeline reproduces the strategy of \textcite{vidalgorene2026paris}, where
Gemini~3 Flash outperformed open-weight VLMs (Qwen2.5-VL, Qwen3-VL) on Armenian text (6.3\,\% CER
against 33--37\,\%); it is the model retained here.} guided by six annotated pages given as
in-context examples, structures each page into records and fields from the OCR text and the page
image, read at high resolution so that the small editorial signs remain legible, and returns one
tagged block per record. A deterministic parser then converts these blocks to JSON
(Appendix~\ref{app:json}), joining records that run across pages and normalising lacuna dots and
line divisions.
Automatic checks finally flag the records to be proofread first: unbalanced or empty brackets,
numbering gaps, and failures of the VLM such as repeated character sequences or the copy of an
in-context example instead of the page. The final extracted corpus
comprises 7,592 records from 2,305 pages, 7,347 transcriptions (about 900,000 characters, 11,479
bracketed editorial spans) and 5,458 detected drawings.

\subsection{Editorial signs}

Editorial signs are the main difficulty of the transcriptions (Table~\ref{tab:signs}). Their use
varies between volumes, each volume's table of conventional signs remaining authoritative: in
volume V, \sig{⎣ ⎦} encloses letters omitted or abbreviated by the engraver and [\,] superfluous
letters on the stone \cite{barkhudaryan1982}, whereas elsewhere square brackets usually enclose
letters supplied by the editor; volume IX marks line divisions by line breaks rather than slashes
\cite{barkhudaryan2012}. Some signs are also rare: square brackets occur in fewer than 4\,\% of
the machine transcriptions, which limits both learning and evaluation. The CIAE reproduces the signs as
printed, normalising only the length of dot runs and the vertical bar that some volumes use for
line divisions. The platform describes them in terms that hold across volumes, ignores them in
search, so that \hy{ՍՈՒՐԲ} also finds \hy{Ս}[\hy{ՈՒՐ}]\hy{Բ}, and rejects any correction whose
brackets do not balance.

\begin{table}[t]
\centering\small
\begin{tabular}{@{}lp{7.6cm}rr@{}}
    \toprule
    Sign & Meaning & Occ. & Error rate \\
    \midrule
    \sig{⎡ ⎤} & text restored by the editor & 134 & 100.7\,\% \\
    \sig{⎣ ⎦} & editor's correction or addition (omission, abbreviation) & 68 & 244.1\,\% \\
        {[\,]} & square brackets as printed, usually letters supplied by the editor & 4 & 75.0\,\%\textsuperscript{a} \\
    ... & lost text (lacuna) & 195 & 0.0\,\% \\
    / & line division as printed & 165 & 63.0\,\% \\
    (\,) & editor's parenthesis, usually a year converted to AD & 56 & 0.0\,\% \\
    \hy{։Ա։} & letters used as numerals & 79 & 1.3\,\% \\
    \midrule
     & Armenian letters & 4,132 & 2.5\,\% \\
     & All characters & 6,027 & 6.7\,\% \\
    \bottomrule
\end{tabular}
\caption{Editorial signs and error rates of the machine reading on the 39 checked records, computed
as (missed + spurious characters) / occurrences, above 100\,\% when signs are inserted; last row:
CER. \textsuperscript{a}Too few occurrences.}
\label{tab:signs}
\end{table}

\begin{figure}[t]
\centering
\begin{tikzpicture}
    \node[anchor=south west, inner sep=0pt] (img) at (0, 0)
        {\includegraphics[width=0.86\linewidth]{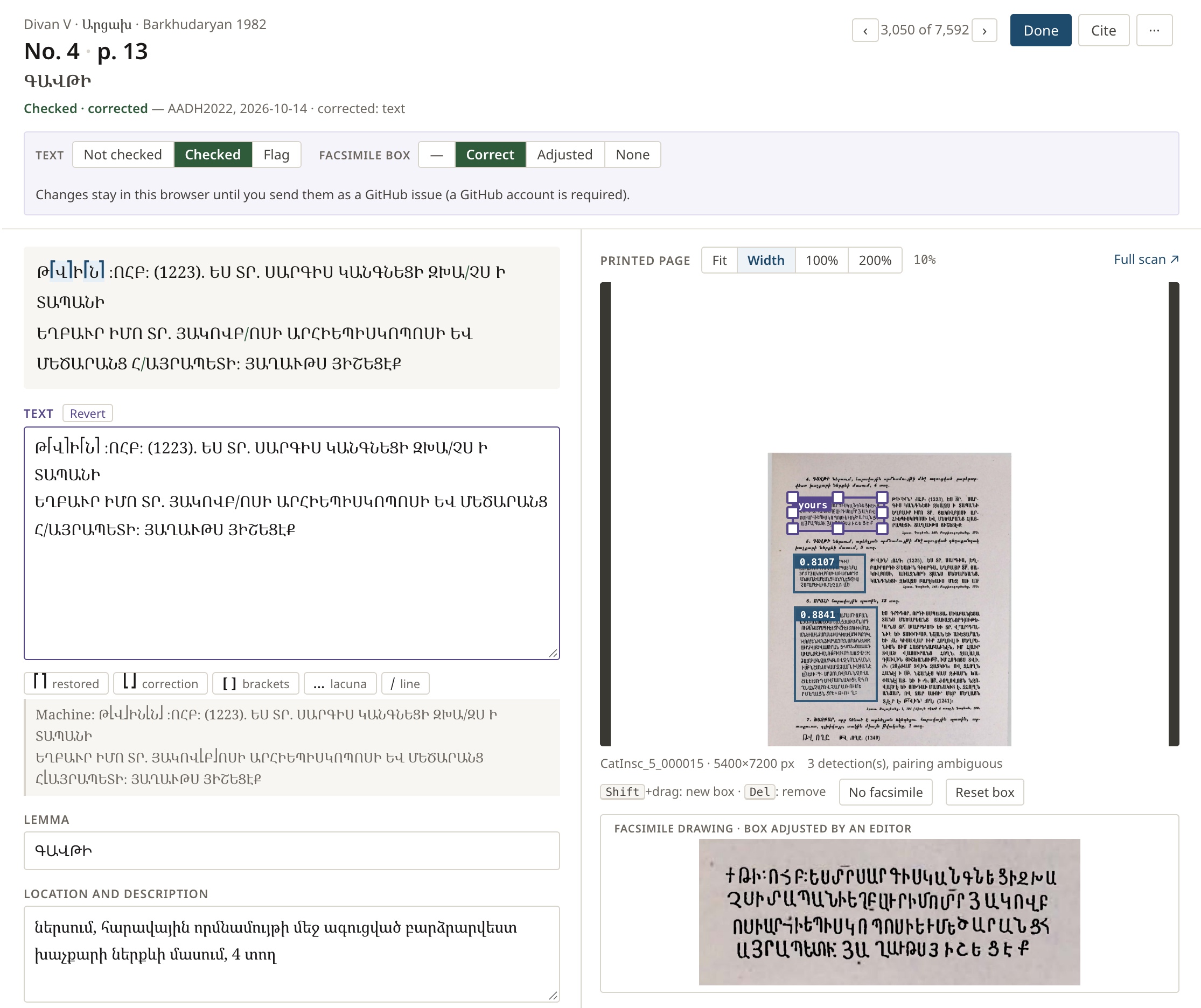}};
    \draw[ciaeFilet, line width=0.6pt] (img.south west) rectangle (img.north east);
    \begin{scope}[x={(img.south east)}, y={(img.north west)}]
        % contributor's name and date, hidden for an anonymous review
        \ifanonymous
            \fill[white] (0.137, 0.881) rectangle (0.258, 0.904);
        \fi
        \callout{0.567}{0.847}{1}
        \callout{0.425}{0.404}{2}
        \callout{0.378}{0.322}{3}
        \callout{0.448}{0.250}{4}
        \callout{0.618}{0.500}{5}
        \callout{0.872}{0.214}{6}
        \callout{0.940}{0.095}{7}
    \end{scope}
\end{tikzpicture}
\caption{Editing \emph{Divan}~V, p.~13, no.~4: (1) text and facsimile-box status;
(2) transcription; (3) editorial-sign buttons; (4) machine reading, kept;
(5) YOLO detections and the editor's box; (6) box tools; (7) resulting facsimile.}
\label{fig:edit}
\end{figure}

\subsection{Online platform and proofreading}

The platform (Figure~\ref{fig:edit}) gives access to the corpus: each record can be searched, read
beside its page scan, cited and exported, with machine values, corrected values or both. It is a
static website whose data are stored in a public GitHub repository, corrections being kept apart
from the machine output. The same interface serves proofreading: specialists edit the fields of a
record, with dedicated controls for the editorial signs, and submit the validated or corrected record
as a pre-filled GitHub issue; an automated workflow checks the submission, then, depending on the
contributor's role, commits it or opens a pull request that an editor reviews against the scan.

Proofreading also covers the drawings: the reviewer accepts, redraws or rejects the facsimile box,
and assigns a drawing to its record where the pairing was ambiguous, a case the pipeline leaves
undecided to avoid false positives.
Work queues list the records flagged by the automatic checks, such as unbalanced brackets, missing
fields and unpaired drawings. Each correction thus also provides annotations: adjusted boxes for the
detector, corrected pages as in-context examples for the VLM, and checked drawing--transcription
pairs for reading models.

\subsection{First results}

The output was first assessed qualitatively: specialists inspected it volume by volume, judged it
usable for search and located most errors in the editorial signs. Rather than defer publication
until a complete ground truth was available, the corpus was opened to collaborative correction,
which provides a quantitative evaluation as it proceeds. As of November 2026, the text of 39 records
has been checked, starting with volume V; on these (6,027 characters), the character error rate is
6.7\,\% with the editorial signs and 2.4\,\% without them (see Table~\ref{tab:signs}). Lacunae,
parentheses and numerals are reproduced almost without error, whereas restoration brackets,
correction brackets and line divisions concentrate the errors. Most errors involve thin strokes:
restoration brackets read as correction brackets (52 of the 101 missed) or omitted (39), line
divisions read as brackets (62 of the 101 missed) or omitted (34), and correction brackets inserted
where the edition has none (45). Across the 7,592 records, the pipeline paired a drawing
automatically in 46\,\% of cases; in 39\,\%, several candidates could not be disambiguated, and in
15\,\%, none was detected.\footnote{Only detections with a confidence above 40\,\% are kept for
display: correcting the remaining cases by hand is faster than reviewing confidence scores, and the
detector can be retrained and run again on the proofread corpus.} Of the 36 facsimile boxes examined
so far, 22 were accepted, 13 redrawn and one rejected.

% ============================================================================
\section{Limitations and perspectives}
% ============================================================================

The evaluation remains preliminary: the 39 records were selected by the proofreaders rather than
sampled, and the error rates will evolve as correction proceeds. The confusions between thin strokes
could be reduced by choosing in-context examples rich in these signs among the corrected records, or
by a post-correction model trained on them. Most facsimiles still await manual confirmation, and the
corpus inherits the readings of the printed edition, errors included, which the platform allows
specialists to correct. Contributions also require a GitHub account, which not all epigraphers have.

An associated project uses the platform to build the first digital corpus of Armenian inscriptions
from Artsakh: volume V is checked first, then the corpus is extended from photographs and drawings,
starting with Martakert. Each inscription receives a diplomatic transcription and an aligned
normalised text that expands abbreviations and regularises spelling, a linguistic markup also
explored for the dialectal variants of Cretan inscriptions \cite{deniz2021}, with metadata following a schema
inspired by EpiDoc. The project builds on a pilot corpus for Shirak (Master's thesis, Inalco,
2022--2023).

The MESROP project (Machine learning for Epigraphic Study and Recognition Of the Armenian Patrimony)
pursues three further aims: training a language model on the corpus, able to
propose readings for lost or worn passages; turning the platform into a hub where photographs of
inscriptions are deposited and linked to their records and to the 3D models of the monuments; and
producing new facsimile drawings, ideally automatically from photographs and 3D models, for the
inscriptions that the \emph{Divan} does not draw and for palaeographic research~\cite{mercier1979}. An EpiDoc export
following \textcite{tamrazyan2026heritage} and the retraining of the detector on corrected boxes are
also planned.

Beyond Armenian, the platform and the process form a proof of concept for similar
corpora: the interface is ready to use and easily adapted to display and correct structured data
produced by VLMs, whatever the language.

% ============================================================================
% NOT COUNTED: acknowledgements, AI disclosure, references
% ============================================================================
\FloatBarrier
\section*{Code and data availability}
\ifanonymous
Links withheld for review.
\else
The code of the platform and the data are available on GitHub; the data are released under
CC BY-SA 4.0 and can also be exported from the platform in CSV, JSON or JSON Lines. Any use should cite the
printed volume as well as the digital record.
\begin{itemize}
    \item Platform (public link): \url{https://epigraphy.calfa.fr}
    \item Platform (code): \url{https://github.com/calfa-co/armenian-inscriptions-website}
    \item Data: \url{https://github.com/calfa-co/armenian-inscriptions-data}
\end{itemize}
\fi

\section*{Acknowledgements}
\ifanonymous
Left blank for review.
\else
The CIAE is led by Calfa in collaboration with Inalco, within the ANR project DALiH (Digitizing
Armenian Linguistic Heritage, ANR-21-CE38-0006). Its work on silver data is pursued with the MESROP
project, funded by the IR* Huma-Num DISTAM+, and the Artsakh corpus with DALiH, Calfa and AGBU. The CIAE rests above all on the
work of compilation and description carried out over nearly six decades by the editors of the
\emph{Divan hay vimagrutʻyan}.
\fi

\section*{Disclosure of generative AI use}
Besides the VLM of the extraction pipeline (Section~\ref{sec:pipeline}), Claude Opus 5.5 (Anthropic)
was used to assist in developing the code of the platform, under the authors' supervision. The same
model was used to check the English of the paper and to help make it more concise. The scientific
content, analyses and conclusions remain the sole responsibility of the authors.

\printbibliography

% ============================================================================
% APPENDIX: figure and caption only
% ============================================================================
\clearpage
\appendix
\section{Data formats}
\label{app:json}

\noindent\begin{minipage}{\linewidth}
% Listings in the monospace font of the URLs; Armenian letters and editorial signs take
% the fonts of the body text through their Unicode definitions (preamble).
\newcommand{\bloc}[1]{\par\vspace{3pt}\noindent #1\par}
\setlength{\topsep}{3pt}\setlength{\partopsep}{0pt}
\fvset{fontsize=\scriptsize, breaklines, frame=single, framesep=4pt}
\bloc{(a) VLM output, tagged text}
\begin{Verbatim}
DESCRIPTION: ներսում, հարավային որմնամույթի մեջ ագուցված բարձրարվեստ խաչքարի ներքևի մասում, 4 տող.
NB_LIGNES: 4
TRANSCRIPTION: Թ⎣Վ⎦ԻՆ⎣Ն⎦ ։ՈՀԲ։ (1223). ԵՍ ՏՐ. ՍԱՐ-
ԳԻՍ ԿԱՆԳՆԵՑԻ ԶԽԱ/ԶՍ Ի ՏԱՊԱՆԻ
ԵՂԲԱՒՐ ԻՄՈ ՏՐ. ՅԱԿՈՎ⎣Բ⎦ՈՍԻ ԱՐ-
ՀԻԵՊԻՍԿՈՊՈՍԻ ԵՎ ՄԵԾԱՐԱՆՑ Հ⎣ԱՅ-
ՐԱՊԵՏԻ։ ՅԱՂԱՒԹՍ ՅԻՇԵՑԷՔ։
REFERENCE: Հրատ. Դադեան, 145։ Բարխուդարեանց, 170.
\end{Verbatim}
\bloc{(b) Record in notices.jsonl (abridged)}
\begin{Verbatim}
{"id": "v5-n4", "volume": "5", "page": "13", "numero": "4", "monument": "ԳԱՎԹԻ",
 "description": "ներսում, հարավային որմնամույթի մեջ … 4 տող",
 "transcription": "Թ⎣Վ⎦ԻՆ⎣Ն⎦ ։ՈՀԲ։ (1223). ԵՍ ՏՐ. ՍԱՐԳԻՍ ԿԱՆԳՆԵՑԻ ԶԽԱ/ԶՍ Ի ՏԱՊԱՆԻ\nԵՂԲԱՒՐ ԻՄՈ ՏՐ. ՅԱԿՈՎ⎣Բ⎦ՈՍԻ ԱՐՀԻԵՊԻՍԿՈՊՈՍԻ ԵՎ ՄԵԾԱՐԱՆՑ Հ⎣ԱՅՐԱՊԵՏԻ։ ՅԱՂԱՒԹՍ ՅԻՇԵՑԷՔ",
 "nb_lignes": "4", "reference": "Հրատ. Դադեան, 145։ Բարխուդարեանց, 170", "note": "",
 "pages": ["CatInsc_5_000015"], "page_geometry": {"CatInsc_5_000015": {"width": 5400, "height": 7200}},
 "yolo_boxes": [{"box": [1752, 2448, 2660, 2798], "conf": 0.7969, "angle": 0},
                {"box": [1764, 3000, 2470, 3395], "conf": 0.8107, "angle": 0},
                {"box": [1771, 3518, 2588, 4445], "conf": 0.8841, "angle": 0}],
 "yolo_box": null, "yolo_pairing": "ambiguous", "raw_text": "DESCRIPTION: … 170."}
\end{Verbatim}
\bloc{(c) Correction in corrections/v5-n4.json}
\begin{Verbatim}
{"crop": {"box": [1752, 2448, 2660, 2798], "page": "CatInsc_5_000015"}, "crop_status": "accepted",
 "fields": {"transcription": "Թ⎡Վ⎤Ի⎡Ն⎤ ։ՈՀԲ։ (1223). ԵՍ ՏՐ. ՍԱՐԳԻՍ ԿԱՆԳՆԵՑԻ ԶԽԱ/ՉՍ Ի ՏԱՊԱՆԻ\nԵՂԲԱՒՐ ԻՄՈ ՏՐ. ՅԱԿՈՎԲ/ՈՍԻ ԱՐՀԻԵՊԻՍԿՈՊՈՍԻ ԵՎ ՄԵԾԱՐԱՆՑ Հ/ԱՅՐԱՊԵՏԻ։ ՅԱՂԱՒԹՍ ՅԻՇԵՑԷՔ"},
 "id": "v5-n4", "review_status": "reviewed", "source_issue": 36,
 "updated_at": "2026-10-14T04:50:46+00:00", "updated_by": "AADH2022"}
\end{Verbatim}
\captionof{figure}{Record v5-n4 (Figure~\ref{fig:edit}): (a)~tagged output of the VLM; (b)~record
in notices.jsonl, abridged; (c)~correction in corrections/v5-n4.json.}
\label{fig:json}
\end{minipage}

\end{document}